# Data-Adaptive Low-Rank Sparse Subspace Clustering

Ivica Kopriva
Division of Computing and Data Science, Ruđer Bošković Institute, Zagreb, Croatia

E-mail: ikopriva@irb.hr

*Abstract*—**Low-rank sparse subspace clustering (LRSSC) algorithms built on self-expressive model effectively capture both the global and local structure of the data. However, existing solutions, primarily based on proximal operators associated with $S_p / \ell_p$, $p\in\{0, 1/2, 2/3, 1\}$, norms are not data-adaptive. In this work, we propose an LRSSC algorithm incorporating a data-adaptive surrogate for the $S_0 / \ell_0$ quasi-norm. We provide a numerical solution for the corresponding proximal operator in cases where an analytical expression is unavailable. The proposed LRSSC algorithm is formulated within the proximal mapping framework, and we present theoretical proof of its global convergence toward a stationary point. We evaluate the performance of the proposed method on three well known datasets, comparing it against LRSSC algorithms constrained by $S_p / \ell_p$, $p\in\{0, 1/2, 2/3, 1\}$, norms.**

***Index Terms—subspace clustering, $\ell_0$ surrogate, proximal operator, data adaptability.***

## I. Introduction

Clustering high-dimensional data into disjoint, homogenous groups is a fundamental problem in data analysis [1] [2]. This problem is highly relevant to various applications, including image segmentation [3] and data mining [4]. In many cases data exhibit fewer degrees of freedom than their ambient domain [5]. This motivates the identification of low-dimensional structures within high-dimensional spaces [2], where data points are assumed to lie close to the union of multiple low-dimensional subspaces [6]-[9]. By leveraging this subspace structure, each data point can be expressed as a linear combination of other points from the same group (subspace), leading to self-expressive model [6]-[9]. Under suitable constraints, optimization procedures yield a representation matrix with a block-diagonal structure [6]-[9], a property that aligns intuitively with spectral clustering methods [9]. Since low-rank SC [6] captures global characteristics of data, while sparse SC [7] captures local characteristics, it is natural to combine low-rank and sparsity constraints to better represent both global and local characteristics. This leads to the development of low-rank sparse SC (LRSSC) [10] [8]. Once the representation matrix is learned, spectral clustering is applied to the estimated graph Laplacian to cluster data points based on the subspaces they originate from [11].

Sparsity is typically measured using the $\ell_0$ quasi-norm, which counts the number of nonzero entries in the representation matrix [8]. This leads to computationally infeasible combinatorial optimization problems that are also highly sensitive to noise. To address this, surrogate functions for the $\ell_0$ quasi-norm have been introduced, as seen in [12]. One common approach involves using $\ell_p$ norms for $p$>0. However, closed-form expressions for corresponding proximal operators exist only for specific values: $p$=1/2 [13], $p$=2/3 [14], and $p$=1 [15]. The proximal operator associate with the $\ell_1$ norm, the soft thresholding operator is known to be biased as it over-penalizes nonzero entries [16]. The over-penalization effect deteriorates as $p\to 0$, see Figure 1. Consequently, optimization with $p$=1/2 or $p$=2/3 is preferred [17]. To bridge the gap in the interval 0<$p$<1, numerous non-convex surrogates of the $\ell_0$ quasi-norm have been proposed (see, for example, Table 1 in [12]). However, existing surrogate functions generally lack adaptability to data.

Motivated by the aforementioned reasons, we propose the following surrogate function for the $\ell_0$ quasi-norm in the LRSSC problem:

$$h_{\delta,n}(x) = 1 - \exp^{-\delta|x|^n} \; n \geq 1, \delta > 0 . \tag{1}$$

For $n$=1 function (1) was proposed in [12], surrogate function no. 3, as a weakly convex surrogate for the $\ell_0$ quasi-norm. For $n$=1, an analytical expression for the corresponding proximal operator is provided in [19]. Because $\lim_{|x|\to\infty} h_{\delta,n}(x) = 1$, the corresponding proximal operator is unbiased. For $n$>1, no closed-form solution exists for the proximal operator. To address this, we derive a numerical method for its computation.

In summary, we propose a general proximal mapping-based approach [20] for data-adaptive LRSSC incorporating $h_{\delta,n}(x)$ penalty to enforce both low-rankness and sparsity. We provide a theoretical proof of the global convergence of the corresponding optimization sequence toward a stationary point. To demonstrate effectiveness of the proposed LRSSC method, we evaluate it on three well-known datasets representing digits, faces and objects comparing its performance against LRSSC models constrained with $S_p / \ell_p$, $p\in\{0, 1/2, 2/3, 1\}$. We illustrate in Figure 1 the penalty functions (left) and the corresponding proximal operators (right). Here the penalty $h_{\delta,n}(x)$ in (1) and corresponding proximal operator are dataset adapted, see Section IV. The MATLAB implementation of the proposed LRSSC algorithm is available at https://github.com/ikopriva/DALRSSC.

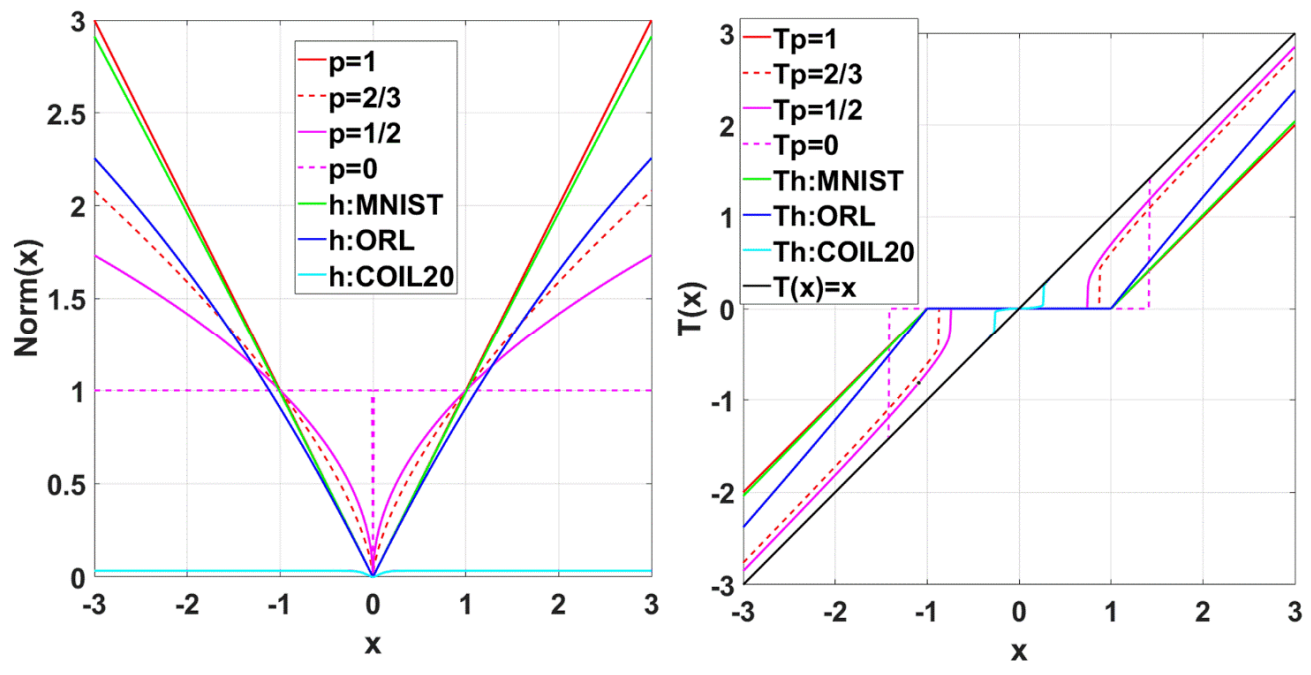

Fig. 1. Penalty functions (left) and corresponding proximal operators (right). Penalty function $h_{\delta,n}(x)$ and proximal operator $T_{h_{\delta,n}}(x,\lambda)$ are data adapted.

## II. Background And Relataed Work

### A. Proximal Operators and Proximal Mappings

The proximal operator of a closed, proper, lower semi-continuous function $h:\mathbb{R}\to\mathbb{R}\cup\infty$ is defined as [21] [22]:

$$prox_h^{\lambda}(x)=\arg\min_u\left(\lambda h(u)+\tfrac{1}{2}\|u-x\|^2\right) \tag{2}$$

Additionally, we introduce the proximity average operator of nonconvex and possibly non-smooth functions $g:\mathbb{R}\to\mathbb{R}\cup\infty$ and $f:\mathbb{R}\to\mathbb{R}\cup\infty$, which are both closed and proper. According to Proposition 1 and Definition 1 in [20], there exists a function that is closed, proper and bounded below such that:

$$\begin{aligned}&prox_{f+g}^{\mu}(x)=\arg\min_u\left(\lambda f(u)+\tau g(u)+\frac{\mu}{2}\|u-x\|^2\right)\\&=\lambda prox_f^{\mu}(x)+\tau prox_g^{\mu}(x)\ \text{ s.t. } \lambda+\tau=1.\end{aligned} \tag{3}$$

To simplify notation we use $T_h(x,\lambda)\triangleq prox_h^{\lambda}(x)$. The proximal operators $T_{\ell_0}(\circ,\lambda)$ and $T_{\ell_1}(\circ,\lambda)$ correspond to the well know hard- and soft-thresholding operators, respectively. These, along with $T_{\ell_{1/2}}(\circ,\lambda)$ and $T_{\ell_{2/3}}(\circ,\lambda)$ have closed-form expressions at $x\in\mathbb{R}$. As previously mentioned, $T_{h_{\delta,1}}(\circ,\lambda)$ has a closed-form expression, which is provided in [19].

### B. Low-Rank Sparse Subspace Clustering

Assuming a normally distributed error term, the LRSSC problem based on the self-expressive property is formulated as:

$$\begin{aligned}&\min_{\mathbf{C}}\tfrac{1}{2}\|\mathbf{X}-\mathbf{XC}\|_F^2+\lambda f(\mathbf{C})+\tau g(\mathbf{C})\\&\text{s.t. } \operatorname{diag}(\mathbf{C})=\mathbf{0}\ \text{ and }\ \lambda+\tau=1.\end{aligned} \tag{4}$$

In (4), $\mathbf{X}\in\mathbb{R}^{D\times N}$ represents dataset consisting of $N$ samples in $D$-dimensional ambient space, and $\mathbf{C}\in\mathbb{R}^{N\times N}$ is self-representation matrix in the frame $\mathbf{X}$. The functions $f$ and $g$ impose low-rank and sparsity constraints on $\mathbf{C}$, respectively, to capture both the global and local structures of the data. The constraint diag($\mathbf{C}$)=$\mathbf{0}$ ensures that data points are not represented by themselves. The additional constraint $\lambda+\tau=1$ arises from the proximal average technique (3), which is used to solve (4). For $S_p/\ell_p$ constrains with $p\in\{0, 1/2, 2/3, 1\}$, we define:

$$f(\mathbf{C})=\|\mathbf{C}\|_{S_p}=\|\boldsymbol{\sigma}(\mathbf{C})\|_{\ell_p},\ g(\mathbf{C})=\|\mathbf{C}\|_{\ell_p} \tag{5}$$

where $\boldsymbol{\sigma}(\mathbf{C})\in\mathbb{R}_{0+}^{N}$ denotes the vector of singular values of $\mathbf{C}$. By introducing an auxiliary variable $\mathbf{J}$, the augmented Lagrangian for (4) is formulated as:

$$\begin{aligned}&\mathcal{L}_{\mu}(\mathbf{J},\mathbf{C},\boldsymbol{\Lambda})=\tfrac{1}{2}\|\mathbf{X}-\mathbf{XJ}\|_F^2+\lambda f(\mathbf{C})+\tau g(\mathbf{C})+\\&\tfrac{\mu}{2}\|\mathbf{J}-\mathbf{C}-\operatorname{diag}(\mathbf{C})\|_F^2+\langle\boldsymbol{\Lambda},\mathbf{J}-\mathbf{C}-\operatorname{diag}(\mathbf{C})\rangle\end{aligned} \tag{6}$$

where $\boldsymbol{\Lambda}\in\mathbb{R}^{N\times N}$ represents the Lagrange multipliers, and $\mu>0$ is a penalty constant. Using the alternating direction method of multipliers (ADMM) [23], we update $\mathbf{J}$ at the iteration $k$+1 as:

$$\mathbf{J}^{k+1}=\left[\mathbf{X}^T\mathbf{X}+\mu^k\mathbf{I}\right]^{-1}\left[\mathbf{X}^T\mathbf{X}+\mu^k\mathbf{C}^k-\boldsymbol{\Lambda}^k\right] \tag{7}$$

By setting $\lambda\neq0$ and $\tau=0$, we obtain the update of $\mathbf{C}$ under the low-rank constraint as:

$$\begin{aligned}&\mathbf{C}_f^{k+1}=\mathbf{U}T_f(\boldsymbol{\Sigma},\lambda/\mu^k)\mathbf{V}^{\mathrm{T}}\quad\text{s.t. } \mathbf{J}^{k+1}+\frac{\boldsymbol{\Lambda}^k}{\mu^k}=\mathbf{U\Sigma V}^{\mathrm{T}}\\&\mathbf{C}_f^{k+1}\leftarrow\mathbf{C}_f^{k+1}-\operatorname{diag}\left(\mathbf{C}_f^{k+1}\right)\end{aligned}. \tag{8}$$

Similarly, by setting $\lambda=0$ and $\tau\neq0$, we obtain the update of $\mathbf{C}$ under the sparsity constraint as:

$$\begin{aligned}&\mathbf{C}_g^{k+1}=T_g(\mathbf{J}^{k+1}+\frac{\boldsymbol{\Lambda}^k}{\mu^k},\tau/\mu^k)\\&\mathbf{C}_g^{k+1}\leftarrow\mathbf{C}_g^{k+1}-\operatorname{diag}\left(\mathbf{C}_g^{k+1}\right)\end{aligned}. \tag{9}$$

By applying the proximal average (3), the update for the representation matrix $\mathbf{C}$ is given by:

$$\mathbf{C}^{k+1}\approx\lambda\mathbf{C}_f^{k+1}+\tau\mathbf{C}_g^{k+1}\quad\text{s.t. } \lambda+\tau=1. \tag{10}$$

The Lagrange multipliers are updated as:

$$\boldsymbol{\Lambda}^{k+1}=\boldsymbol{\Lambda}^k+\mu^k\left(\mathbf{J}^{k+1}-\mathbf{C}^{k+1}\right). \tag{11}$$

The penalty constant is updated as:

$$\mu^{k+1}=\min\left(\rho\mu^k,\mu^{\max}\right) \tag{12}$$

where $\rho>1$ and $\mu^{\max}=10^6$. The algorithm terminates when:

$$\left\|\mathbf{J}^{k+1}-\mathbf{C}^{k+1}\right\|_{\infty}\leq\varepsilon\ \text{ or }\ k>k^{\max}. \tag{13}$$

In our implementation, we set $\varepsilon=10^{-4}$ and $k^{\max}=100$. To further increase robustness against errors, we leverage the mathematically tractable intra-subspace projection dominance (IPD) property [25] to threhsold the learned representation matrix $\mathbf{C}$:

$$\mathbf{C}\leftarrow\lceil\mathbf{C}\rceil_d \tag{14}$$

where the operator $\lceil\circ\rceil_d$ is applied column-wise, retaining the $d$ largest coefficients (in absolute value) and setting the remaining entries to zero. In (14), $d$ represents the *a priori* known subspace dimension. According to the literature, for the MNIST dataset, $d$=12 [26], while for the ORL and COIL20 datasets, $d$=9 [27]. Using $\mathbf{C}$, the data affinity matrix is computed as $\mathbf{A}=\left(|\mathbf{C}|+|\mathbf{C}|^{\mathrm{T}}\right)/2$. Bases on $\mathbf{A}$, the normalized graph Laplacian matrix is computed as $\mathbf{L}=\mathbf{I}-(\mathbf{D})^{-1/2}\mathbf{A}(\mathbf{D})^{-1/2}$, where $\mathbf{D}$ is the diagonal degree matrix with elements $\mathbf{D}_{ii}=\sum_{j=1}^{N}\mathbf{A}_{ij}$. The spectral clustering algorithm [11], is then applied to $\mathbf{L}$ to assign the cluster labels to data points: $\mathbf{F}\in\mathbb{N}_0^{N\times C}$, where $C$ represents the number of clusters.

## III. Data Adaptable Proximal Operator

To obtain the data adaptable version of the LRSSC algorithm (4)-(14) we select:

$$f(\mathbf{C})=h_{\delta,n}(\boldsymbol{\sigma}(\mathbf{C}))=\sum_{i=1}^{N}h_{\delta,n}(\sigma_i(\mathbf{C}))$$

$$g(\mathbf{C})=h_{\delta,n}(\mathbf{C})=\sum_{i,j=1}^{N}h_{\delta,n}(\mathbf{C}_{ij}) \tag{15}$$

where $h_{\delta,n}(x)$ is given in (1). To implement the LRSSC algorithm, we need to compute the proximal operator $T_{h_{\delta,n}}$ to be used in (8) and (9). For $n$=1, a closed-form expression is provided in [19]. We can rewrite (2) as:

$$\begin{aligned}prox_{h_{\delta,n}}^{\lambda}(x)&=\arg\min_u\left(\lambda h_{\delta,n}(u)+\tfrac{1}{2}\|u-x\|^2\right)\\&=\arg\min_u L(u,x)\end{aligned}. \tag{16}$$

It follows that:

$$L(u,-x)=\tfrac{1}{2}(u+x)^2+\lambda h_{\delta,n}(u) \tag{17}$$

and

$$L(-u,x)=\tfrac{1}{2}(u+x)^2+\lambda h_{\delta,n}(u)=L(u,-x). \tag{18}$$

Based on (18), it follows that:

$$u^*=\arg\min_u L(u,x)\ \text{ for } x\geq 0 \tag{19}$$

and:

$$u^*=-\arg\min_u L(-u,x)\ \text{ for } x<0. \tag{20}$$

Thus, we obtain:

$$\begin{aligned}u^*&=\operatorname{sign}(x)\arg\min_u\left\{\tfrac{1}{2}(\operatorname{sign}(x)u-x)^2+\lambda h_{\delta,n}(u)\right\}\\&=\operatorname{sign}(x)\arg\min_u\left\{\tfrac{1}{2}(u-|x|)^2+\lambda h_{\delta,n}(u)\right\}\end{aligned}. \tag{21}$$

It follows from (21) that $u^*\geq 0$, [19]. If $u^*<0$, then $\hat{u}=-u^*$ contradicts optimality of $u^*$. For $u^*<0$ and $\hat{u}=-u^*>0$ it applies $(\hat{u}-|x|)^2<(u^*-|x|)^2$. However, we have $(\hat{u}-|x|)^2=(u^*+|x|)^2>(u^*-|x|)^2$. This leads to a contradiction and, thus, we conclude that $u^*\geq 0$. Therefore, (21) becomes:

$$u^*=\operatorname{sign}(x)\arg\min_u\left\{\tfrac{1}{2}(u-|x|)^2+\lambda h_{\delta,n}(u)\right\}\ u\geq 0. \tag{22}$$

Let us define:

$$\hat{L}(u,x)=\tfrac{1}{2}(u-|x|)^2+\lambda h_{\delta,n}(u),\ u\geq 0. \tag{23}$$

Next, we introduce the derivative of (23) as:

$$\hat{L}'(u,x)=\frac{\partial\hat{L}(u,x)}{\partial u}=u-|x|+\lambda\sigma n u^{n-1}e^{-\delta u^n}\quad u\geq 0. \tag{24}$$

The extremal points of $\hat{L}(u,x)$ in (23) are the zeros of $\hat{L}'(u,x)$ in (24). According to Theorem 1 in [24], the zeros of $\hat{L}'(u,x)$ lie within the interval [$\alpha$, $\beta$], with the endpoints having opposite signs, i.e., $\hat{L}'(u,\alpha)\hat{L}'(u,\beta)<0$. We now introduce the second derivative of (23):

$$\hat{L}''(u,x)=\frac{\partial\hat{L}'(u,x)}{\partial u}=\lambda\sigma n u^{n-2}e^{-\delta u^n}\left(n-1-\delta n u^n\right),\ u\geq 0. \tag{25}$$

The extremes of $\hat{L}'(u,x)$ correspond to the zeros of $\hat{L}''(u,x)$. Through the analysis of (25) we can see that the extremes of $\hat{L}'(u,x)$ occur at three values of $u$:

$$\lambda\delta n u^{n-2}=0\Rightarrow u_1=0. \tag{26a}$$

$$n-1-\delta n u^n=0\Rightarrow u_2=\sqrt[n]{\frac{n-1}{\delta n}} \tag{26b}$$

$$e^{-\delta u^n}\approx 0\Rightarrow e^{-\delta u^n}=e^{-d}\,(d\gg 0)\Rightarrow u_3=\sqrt[n]{\frac{d}{\delta}}\to\infty \tag{26c}$$

Since $\hat{L}'(u_1,x)=-|x|$, the sign of $\hat{L}'(u_1,x)$ is negative. Since $\lim_{u_3\to\infty}\hat{L}'(u_3,x)=\infty$, the sign of $\hat{L}'(u_3,x)$ is positive. Thus, sign of $\hat{L}'(u_2,x)$ is critical for locating the endpoints of the interval containing zero of $\hat{L}'(\xi,x)$ such that $\xi=\arg\min_u\hat{L}(u,x)$. By substituting (26b) into (24) we obtain:

$$\hat{L}(u_2,x)=\sqrt[n]{\frac{n-1}{\delta n}}+\lambda\delta n\left(\frac{n-1}{\delta n}\right)^{\frac{n-1}{n}}e^{-\delta\left(\sqrt[n]{\frac{n-1}{\delta n}}\right)^n}-|x|. \tag{27}$$

The sign of $\hat{L}((u_2),x)$ depends on $n$, $\delta$, and $x$ and can be either negative or positive. To find the minimum of $\hat{L}(u,x), u\geq 0$ in (23), we use the method of nonlinear least squares, implemented using the Matlab function `lsqnonlin` with the trust-region algorithm. We solve (23) twice. The first time we set the initial condition to $u_0=u_2/2$, which yields the optimal solution $u_{\min}^1$. The second time we set the initial condition to $u_0=|x|$, which yields the optimal solution $u_{\min}^2$. We then evaluate (23) at $\hat{L}\left(u_{\min}^1, x\right)$ and $\hat{L}\left(u_{\min}^2, x\right)$, and obtain the minimum of (23) as:

$$\begin{aligned} &\text{If } \hat{L}\left(u_{\min}^1, x\right) < \hat{L}\left(u_{\min}^2, x\right) \text{ then } u_{\min} = u_{\min}^1 \\ &\text{else } u_{\min} = u_{\min}^2 \text{ end.} \end{aligned} \tag{28}$$

Combining (16), (21) and (28) we finally obtain:

$$T_{h_{\delta,n}}\left(x, \lambda\right) = sign\left(x\right) u_{\min}. \tag{29}$$

**Remark 1**. The assertion is made in [19] for problem (2) with $h(x) = h_{\delta,1}(x)$. The assertion applies to any value of $n$. The regularization constant $\lambda$ can be a function of $\delta$, i.e., $\lambda(\delta) = 1/\delta$. That yields (2) with the regularization function $\left(\frac{1}{\delta}\right) h_{\delta,n}(x)$. Then, $\lim_{\delta \to 0} T_{\frac{1}{\delta} h_{\delta,1}}(x) = \operatorname{sign}(x) \max\left(|x| - 1, 0\right)$, which is the soft-thresholding operator. Also, for $x \neq 0$, $\lim_{\delta \to 0} h_{\delta,1}(x) \approx \delta$. Thus, $T_{h_{\delta,1}}(x)$ behaves like a hard thresholding operator with the thresholding region controlled by $\delta$.

**Theorem 1**. Let $Y^k = \left\{\left(\mathbf{J}^k, \mathbf{C}^k, \mathbf{\Lambda}^k\right)\right\}_{k=1}^{\infty}$ be a sequence generated by the LRSSC algorithm (4)-(13). Then, for sufficiently large $\mu$, the algorithm converges globally.

*Proof.* The proof is similar to that of Theorem 2 in [8]. To prove global convergence, we need to verify the fulfillment of assumptions A1-A5 established in [28]. The only difference compared to [8] is the regularization functions in (15). Since these are real analytic functions, it is straightforward to confirm the fulfillment of assumptions A1-A5. Additionally, $\mathcal{L}_\mu\left(\mathbf{C}, \mathbf{J}, \mathbf{\Lambda}\right)$ in (6) is also real-analytic, which makes it a Kurdyka-Łojasiewicz function [29]. Therefore, by [28], the algorithm (4)-(13) converges globally to the unique stationary point of $\mathcal{L}_\mu\left(\mathbf{C}, \mathbf{J}, \mathbf{\Lambda}\right)$. €

## IV. Experimental results

We validated the proposed data-adaptive LRSSC algorithm with low-rank and sparsity regularization functions in (5) and (15) on three well-known datasets: MNIST [26], COIL20 [27] and ORL [30]. The MNIST dataset contains digit images grouped into $C$=10 clusters, with 1000 samples per cluster. The ORL dataset contains face images grouped into $C$=40 clusters, with 10 samples per cluster. The COIL20 dataset contains images of objects grouped into $C$=20 clusters, with 72 samples per cluster. We tuned hyperparameters on held-out datasets and estimated clustering performance on 100 random in-sample and out-of-sample partitions with sizes of 50/50, 7/3 and 26/26, respectively. For both $\frac{1}{\delta} h_{\delta,n}(x)$ and $h_{\delta,n}(x)$, parameters were tuned across the range $\delta \in$ {1/50 1/5 2/5 7/10 10:10:50} and $n \in$ {1:0.1:2}. For all LRSSC algorithms the regularization constant was tuned in the range $\lambda \in$ {0:0.1:1}, $\rho$=3 for all datasets, and $\mu^0$ was tuned in the range {2:0.5:4}. The performance metrics used were accuracy (ACC), normalized mutual information (NMI), and $F_1$ score. They take values in the range [0, 1], where 0 indicates the worst performance and 1 indicates the best. Results are presented in table 1 in terms of mean and standard deviation (in subscript). An asterisk (*) implies that, according to the non-parametric Wilcoxon sum-rank test with $p$<0.05, there is no statistically significant difference with respect to the LRSSC regularized with (15). The figure 1 shows the regularization functions (5) and (15), as well as the corresponding proximal operators used in the LRSSC algorithms. For MNIST, ORL and COIL20 datasets proposed approach selected the following penalties in respective order: $50h_{\frac{1}{50},1}$, $5h_{\frac{1}{5},1}$ and $\frac{1}{30} h_{30,1.5}$. As can be seen in table 1, it adapts to data and achieves performance comparable to the best $S_p/\ell_p$ regularized LRSSC algorithm. However, due to numerical computation of proximal operator, computational complexity of proposed approach is up to two orders of magnitude greater than the one of $S_p/\ell_p$ regularized LRSSC algorithms. Therefore, its use needs to be justified by the specific application.

TABLE I: Clustering performance on Mnist, Orl and Coil20 datasets.

| Algorithm | ACC [%] | NMI [%] | $F_1$ [%] |
|---|---|---|---|
| | | MNIST | |
| $50h_{\frac{1}{50},1}$ | $61.8_{4.3}/61.4_{4.3}$ | $64.6_{2.9}/64.2_{2.7}$ | $53.9_{3.7}/53.3_{3.6}$ |
| $S_1/\ell_1$ | $59.5_{3.8}/59.5_{3.8}$ | $62.7_{2.9}/62.8_{2.9}$ | $62.8_{2.9}/51.6_{3.4}$ |
| $S_{2/3}/\ell_{2/3}$ | $^*62.1_{4.1}/^*62.2_{4.2}$ | $^*63.8_{2.9}/^*64.0_{3.0}$ | $63.8_{2.9}/^*53.5_{3.8}$ |
| $S_{1/2}/\ell_{1/2}$ | $58.3_{4.3}/58.6_{4.1}$ | $61.5_{3.0}/62.00_{2.7}$ | $50.1_{3.6}/50.6_{3.5}$ |
| $S_0/\ell_0$ | $58.4_{4.5}/58.8_{4.6}$ | $61.7_{3.3}/62.3_{3.1}$ | $50.3_{4.0}/50.9_{3.9}$ |
| | | ORL | |
| $5h_{\frac{1}{5},1}$ | $75.9_{2.5}/73.5_{3.0}$ | $86.7_{1.1}/88.4_{1.4}$ | $63.8_{3.0}/53.8_{4.9}$ |
| $S_1/\ell_1$ | $76.6_{2.5}/74.0_{3.0}$ | $87.17_{1.3}/^*88.7_{1.3}$ | $87.2_{1.2}/55.2_{4.5}$ |
| $S_{2/3}/\ell_{2/3}$ | $^*75.3_{2.5}/^*73.0_{3.0}$ | $^*86.8_{1.2}/^*88.2_{1.2}$ | $86.8_{1.2}/{*}53.4_{4.3}$ |
| $S_{1/2}/\ell_{1/2}$ | $^*76.0_{2.4}/^*73.5_{2.7}$ | $87.2_{1.2}/^*88.5_{1.2}$ | $^*64.2_{3.0}/^*54.1_{4.5}$ |
| $S_0/\ell_0$ | $69.1_{3.0}/67.1_{3.0}$ | $85.0_{1.3}/86.1_{1.4}$ | $56.5_{3.4}/45.9_{4.5}$ |
| | | COIL20 | |
| $\frac{1}{30} h_{30,1.5}$ | $81.0_{2.1}/81.2_{2.0}$ | $89.1_{1.0}/89.0_{1.0}$ | $77.0_{2.0}/76.3_{1.9}$ |
| $S_1/\ell_1$ | $76.5_{2.2}/77.0_{2.1}$ | $86.0_{1.1}/86.0_{1.0}$ | $86.0_{1.1}/71.4_{2.3}$ |
| $S_{2/3}/\ell_{2/3}$ | $79.8_{2.2}/79.7_{2.1}$ | $90.5_{1.0}/90.3_{1.1}$ | $90.5_{1.0}/75.3_{1.8}$ |
| $S_{1/2}/\ell_{1/2}$ | $78.1_{2.9}/78.3_{2.9}$ | $^*88.7_{1.0}/88.8_{1.0}$ | $74.2_{2.4}/75.4_{2.4}$ |
| $S_0/\ell_0$ | $79.0_{2.4}/79.1_{2.4}$ | $88.7_{1.1}/88.5_{1.2}$ | $75.4_{2.2}/74.5_{2.2}$ |

## V. Conclusion

In this paper, we proposed a data-adaptive low-rank sparse subspace clustering (LRSSC) algorithm with a flexible low-rank and sparsity regularization function. We provide a numerical method for calculating the corresponding proximal operator and proved the global convergence of the proposed algorithm. Proposed LRSSC algorithm demonstrated the ability to adapt to datasets and to achieve comparable or better performance than best $S_p/\ell_p$ regularized LRSSC algorithm. Our future work will be directed toward computationally more efficient numerical solution.

## Acknowledgment

This work was supported by the Croatian Science Foundation grant under the project number HRZZ-IP-2022-10-6403.

## REFERENCES

[1] R. Xu and D. Wunsch, "Survey of clustering algorithms," *IEEE Transactions on Neural Networks*, vol. 16, no. 3, pp. 645-678, 2005.

[2] J. Wight and Y. Ma, High-Dimensional Data Analysis with Low-Dimensional Models - Principles, Computation and Applications, Cambridge University Press, 2022.

[3] L. Wang and C. Pan, "Robust level set image segmentation via a local correntropy-based k-means clustering," *Pattern Recognition*, vol. 47, no. 5, pp. 1917-1925, 2014.

[4] W. Wu, and M. Peng, "A data mining approach combining *k*-means clustering with bagging neural network for short-term wind power forecasting," *IEEE Internet of Things Journal*, vol. 4, no. 4, pp. 979-986, 2017.

[5] M. Udell, and A. Townsend, Why are big data matrices approximately low rank? *SIAM J. Math. Data Sci*., vol. 1, no. 1, pp. 144–160, 2019.

[6] G. Liu, Z. Lin, S. Yan, J. Sun, Y. Yu, and Y. Ma, "Robust recovery of subspace structures by low-rank representation," *IEEE Trans. Patt. Anal. Mach. Intell.*, vol. 35, no. 1, pp. 171–184, 2013.

[7] E. Elhamifar and R. Vidal, "Sparse Subspace Clustering: Algorithm, Theory, and Applications," *IEEE Trans. Patt. Anal. Mach. Intell.*, vol. 35, no. 1, pp. 2765-2781, 2013.

[8] M. Brbić and I. Kopriva, " $\ell_0$ Motivated Low-Rank Sparse Subspace Clustering," *IEEE Trans. Cyber.*, vol. 50, no. 4, pp. 1711-1725, 2020.

[9] C. Lu, J. Feng, Z. Lin, T. Mei, and S. Yan, "Subspace Clustering by Block Diagonal Representation," *IEEE Trans. Patt. Anal. Mach. Intell.*, vol. 41, no. 2, pp. 487-501, 2018.

[10] C.-G. Li and R. Vidal, "A structured sparse plus structured low-rank frame work for subspace clustering and completion," *IEEE Trans. Signal Process.*, vol. 64, no. 24, pp. 6557–6570, Dec. 2016.

[11] U. von Luxburg, "A tutorial on spectral clustering," *Stat. Comput.*, vol. 17, no. 4, pp. 395–416, 2007.

[12] L. Chen, X. Gu, "The Convergence Guarantees of a Non-Convex Approach for Sparse Recovery," *IEEE Trans. Sig. Proc.*, vol. 62, no. 15, pp. 3754-3767, 2014.

[13] Z. Xu, X. Chang, F. Xu, and H. Zhang, "L-1/2 regularization: a thresholding representation theory and a fast solver", *IEEE Trans. Neural Netw. Learn. Syst*., vol. 135, pp. 1013–1027, 2012.

[14] W. Cao, J. Sun, and Z. Xu, "Fast image deconvolution using closed-form thresholding fromulas of $L_q$ ( $q = 2/3, 1/2$ ) regularization," *J. Vis. Commun. Image. Represent.*, vol. 24, no. 1, pp. 31-41, 2013.

[15] F. Chen, L. Shen, and B. W. Suter, "Computing the proximity operator of the $\ell_p$ norm with 0<*p*<1," *IET Signal Process.*, vol. 10, no. 5, pp. 557–565, 2016.

[16] E. J. Candes, M. B. Walkin, S. P. Boyd, "Enhancing Sparsity by reweighted $\ell_1$ minimization," *J. Fourier Anal. and Appl.*, vol. 14, no. 5, pp. 877-905, 2008.

[17] S. Foucart, M. Lai, "Sparsest solution of underdetermined linear systems via $\ell_q$ -minimization for 0<*q*≤1," *Appl. Comput. Harmon. Anal.*, vol. 26, no. 3, pp. 395-407, 2009.

[18] F. Wen, L. Chu, O. Lu, and R. Qiu, "A Survey on Nonconvex Regularization Based Sparse and Low-Rank Recovery in Signal Processing, Statistics, and Machine Learning," *IEEE Access*, vol. 6, pp. 69883-69906, 2018.

[19] M. Malek-Mohammadi, A. Koochakzadeh, M. Babaie-Zadeh, M. Jansson, and C. R. Rojas, "Successive Concave Sparsity Approximation for Compressed Sensing," *IEEE Trans. Sig. Proc.*, vol. 64, no. 21, pp. 5657-5671, 2016.

[20] Y. Yu, X. Zheng, M. Marchetti-Bowick, and E. Xing, "Minimizing nonconvex non-separable functions," in *Proc. 18th Int. Conf. Artif. Intell. Stat.*, vol. 38, 2015, pp. 1107–1115.

[21] J. Bolte, S. Sabach, and M. Teboulle, "Proximal alternating linearized minimization for nonconvex and nonsmooth problems," *Math. Program*., vol. 146, nos. 1–2, pp. 459–494, 2014.

[22] N. Parikh and S. Boyd, "Proximal algorithms," *Foundations and Trends in Optimization*, vol. 1, pp. 123-231, 2013.

[23] S. Boyd, N. Parikh, E. Chu, B. Peleato, and J. Eckstein, "Distributed optimization and statistical learning via the alternating direction method of multipliers," *Found. Trends Mach. Learn.*, vol. 3, no. 1, pp. 1–122, 2011.

[24] B. P. Demidovich, I. A. Maron, "Approximate solutions of algebraic and transcendental equations," Chapter 4 in *Computational Mathematics*, Mir Publishers, Moscow, 4th printing, 1987.

[25] Xi Peng, Zhiqiang Yu, Zhang Yi, and Huajin Tang. Constructing the l2-graph for robust subspace learning and subspace clustering. *IEEE Transactions on Cybernetic*s, vol. 47, no. 4, pp. 1053–1062, 2017.

[26] Y. LeCun, L. Bottou, Y. Bengio, and P. Haffner, "Gradient-based learning applied to document recognition," *Proceedings of The IEEE*, vol. 86,no. 11, pp. 2278–2324, 1998.

[27] S. A. Nene, S. K. Nayar, and H. Murase, "Columbia object image library (coil-100)," Technical Report, CUCS-006-96, Dept. of Computer Science, Columbia Univ, 1996,

[28] Y. Wang, W. Yin, and J. Zeng, "Global convergence of ADMM in nonconvex nonsmooth optimization," *J. Sci. Comput.*, vol. 78, pp. 29-63, 2019.

[29] S. Łojasiewicz, "Une propriété topologique des sous-ensembles analytiques réels," in: *Les Équations aux Dérivées Partielles*, pp. 87–89, Éditions du centre National de la Recherche Scientifique, Paris 1963.

[30] F. S. Samaria and A. C. Harter, "Parameterisation of a stochastic model for human face identification," in *WACV. IEEE*, 1994, pp. 138–142.